\documentclass[letterpaper]{article} % DO NOT CHANGE THIS
\usepackage{aaai2026}  % DO NOT CHANGE THIS
\usepackage{times}  % DO NOT CHANGE THIS
\usepackage{helvet}  % DO NOT CHANGE THIS
\usepackage{courier}  % DO NOT CHANGE THIS
\usepackage[hyphens]{url}  % DO NOT CHANGE THIS
\usepackage{graphicx} % DO NOT CHANGE THIS
\usepackage{natbib}  % DO NOT CHANGE THIS AND DO NOT ADD ANY OPTIONS TO IT
\usepackage{caption} % DO NOT CHANGE THIS AND DO NOT ADD ANY OPTIONS TO IT
\usepackage{pifont}
\usepackage{makecell}
\usepackage{amsfonts}
\usepackage{algorithm}
\usepackage{algorithmic}
\usepackage{multirow} 
\usepackage{booktabs} 
\usepackage{colortbl,xcolor}
\usepackage{newfloat}
\usepackage{listings}
\DeclareCaptionStyle{ruled}{labelfont=normalfont,labelsep=colon,strut=off} % DO NOT CHANGE THIS
\floatstyle{ruled}
\newfloat{listing}{tb}{lst}{}
\floatname{listing}{Listing}
\title{When Privacy Meets Recovery: The Overlooked Half of \\Surrogate-Driven Privacy Preservation for MLLM Editing}
\author{
    Siyuan~Xu\textsuperscript{\rm 1},
    Yibing~Liu\textsuperscript{\rm 1}\thanks{Corresponding author.},
    Peilin~Chen\textsuperscript{\rm 1},
    Yung-Hui~Li\textsuperscript{\rm 2},
    Shiqi~Wang\textsuperscript{\rm 1},
    Sam~Kwong\textsuperscript{\rm 3}
    }
\affiliations{
    \textsuperscript{\rm 1}City University of Hong Kong\\
    \textsuperscript{\rm 2}Hon Hai Research Institute\\
    \textsuperscript{\rm 3}Lingnan University\\
    siyuanxu333@gmail.com, lyibing112@gmail.com, plchen3@cityu.edu.hk, \\yunghui.li@foxconn.com, shiqwang@cityu.edu.hk, samkwong@ln.edu.hk
}
\usepackage{bibentry}
\usepackage{xcolor} 
\usepackage{amsmath}
\usepackage{arydshln}
\begin{document}

\maketitle
\begin{abstract}

Privacy leakage in Multimodal Large Language Models (MLLMs) has long been an intractable problem. Existing studies, though effectively obscure private information 
in MLLMs, often overlook the evaluation of the authenticity and recovery quality of user privacy. To this end, this work uniquely focuses on the critical challenge of how to restore surrogate-driven protected data in diverse MLLM scenarios. 
We first bridge this research gap by contributing the SPPE (Surrogate Privacy Protected Editable) dataset, which includes a wide range of privacy categories and user instructions to simulate real MLLM applications. This dataset offers protected surrogates alongside their various MLLM-edited versions, thus enabling the direct assessment of privacy recovery quality. By formulating privacy recovery as a guided generation task conditioned on complementary multimodal signals, we further introduce a unified approach that reliably reconstructs private content while preserving the fidelity of MLLM-generated edits. 
The experiments on both SPPE and InstructPix2Pix further show that our approach generalizes well across diverse visual content and editing tasks, achieving a strong balance between privacy protection and MLLM usability.

\end{abstract}

% Uncomment the following to link to your code, datasets, an extended version or similar.
% You must keep this block between (not within) the abstract and the main body of the paper.
% \begin{links}
%     \link{Code}{https://aaai.org/example/code}
%     \link{Datasets}{https://aaai.org/example/datasets}
%     \link{Extended version}{https://aaai.org/example/extended-version}
% \end{links}

\section{Introduction}
\begin{figure}[!h]
    \centering
    \includegraphics[width=.85\linewidth]{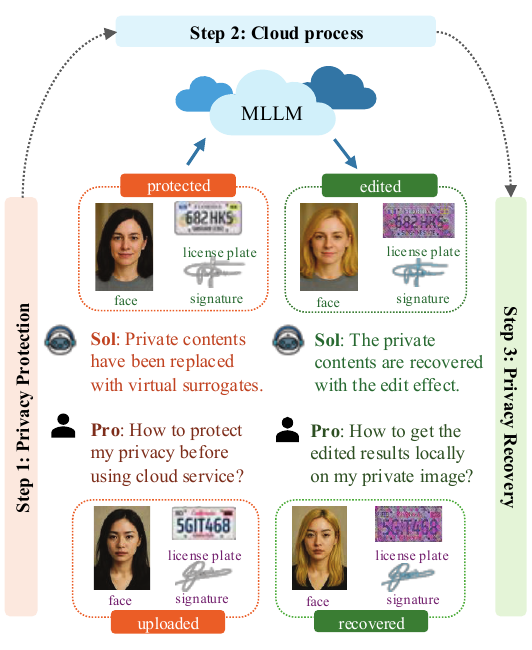}
    \caption{Demonstration of our Edit-Compatible Surrogate-Driven Privacy Protection paradigm. Sensitive regions in the original image are locally replaced with synthetic content to create a surrogate, which is sent to the cloud for editing by MLLMs. The surrogate's edits are locally combined with the original image to produce a privacy-preserving output that faithfully reflects MLLM-intended modifications.}
    \label{fig:pipeline}
\end{figure}
\begin{figure*}[h]
    \centering
    \includegraphics[width=.92\linewidth]{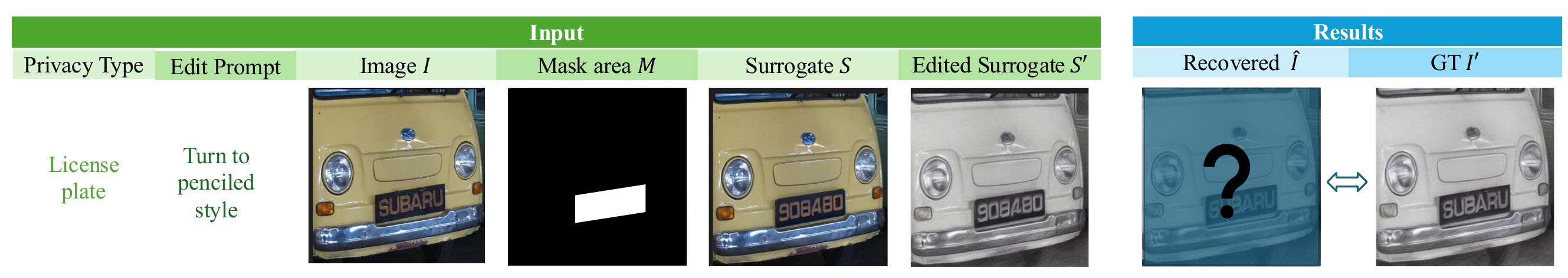}
    \caption{The sensitive category (C) is “license plate,” and the edit prompt is “Turn to penciled style.” The original image $I$ contains the private content “SUBARU,” which is replaced by a synthetic one, “908ABD,” in the surrogate image $S$. However, the MLLM-edited surrogate output $S'$ retains the synthetic plate “908ABD” rather than reflecting the original content, necessitating recovery of the surrogate output to better approximate the edited original image $I'$. }
    \label{fig:sample}
\end{figure*}

The rapid development of Multimodal Large Language Models (MLLMs) has unlocked powerful reasoning over complex visual inputs. However, this strength is often accompanied by serious privacy risks, as diverse and subtle private cues embedded in visual input can be easily captured and exposed. While prior work~\cite{revision} has leveraged textual surrogates (e.g., descriptions or summaries of visual content) to bypass direct image uploads for tasks such as visual question answering, these approaches fall short in scenarios that require direct manipulation of visual inputs, such as image editing. This motivates the use of visual surrogates—synthetic substitutes for sensitive regions—which have demonstrated strong privacy protection through perceptual concealment~\cite{chi}, while retaining high utility for downstream analysis~\cite{hrvispr}. 
However, edits performed on surrogates often deviate from those on the original image, which makes it difficult to preserve the visual-semantic consistency with the original.
To address this, our work focuses on the post-surrogate stage, defining a recovery process that faithfully reconstructs outputs while preserving both the consistency of the intended edits and the integrity of the original content.

As illustrated in Figure~\ref{fig:pipeline}, the Edit-Compatible Surrogate-Driven Privacy Protection paradigm leverages surrogates to overcome the challenges in privacy-preserving editing. This process begins locally, where privacy-sensitive regions are replaced with synthesized surrogates containing no original private information. These surrogates are then edited by the MLLM in the cloud while the MLLM is restricted to interacting solely with the surrogate image. Subsequently, the edited surrogates serve as semantic references to guide the integration of the editing intent with the original image locally. The final result is then locally recovered as an edited version of the original image by simulating the MLLM’s editing behavior based on the transformation observed in the edited surrogate.

To provide a unified standard for this paradigm, we introduce \textbf{SPPE-Bench}. This benchmark is designed to evaluate recovery quality with respect to both MLLM editing fidelity and privacy preservation under surrogate-based protection. The dataset covers a broad range of privacy-sensitive categories and diverse editing instructions, simulating realistic scenarios. Building on our dataset, we propose Surrogate-to-Original Editable Recovery (\textbf{SOER}), a unified framework that reconstructs MLLM-style edits on original images without exposing them to the MLLM. We leverage a Diffusion Transformer (DiT) to jointly model semantic, visual, and spatial cues, guiding the generation process to faithfully reproduce the intended edits while maintaining consistency with the original visual content. Extensive evaluations on the SPPE benchmark and the InstructPix2Pix dataset demonstrate strong performance in edit fidelity, privacy preservation, and generalization.
Our main contributions are summarized as follows:

\begin{itemize}
  \item We center on an under-explored privacy problem in MLLM services, emphasizing the challenge of enabling effective image editing under surrogate-driven privacy constraints.
\item We introduce \textbf{SPPE}, to the best of our knowledge, the first benchmark dataset explicitly designed to evaluate edit fidelity of utilizing MLLMs under privacy-preserving conditions.
\item We propose \textbf{SOER}, a multimodal generative framework that achieves privacy-aware recovery of MLLM outputs across diverse sensitive content and editing instructions using a single model.
\end{itemize}

\section{Related Work}

\subsection{Multimodal Large Language Model}
Multimodal Large Language Models (MLLMs) have shown impressive capabilities in aligning and reasoning over multimodal inputs. Models like LLaVA and MiniGPT-4~\cite{llava, minigpt4} leverage instruction tuning on top of LLaMA~\cite{llama} to enable strong performance in tasks such as VQA, grounded reasoning, and human-AI interaction. To extend MLLMs to image synthesis, GILL~\cite{gill} integrates LLMs with diffusion models, while SEED and SEED-2~\cite{seed, seed2} propose visual tokenizers that align image and text embeddings for coherent generation. Recently, SmartEdit~\cite{smartedit} introduces a Bidirectional Interaction Module to enhance instruction understanding and editing precision. 
Together, these advances highlight the shift toward generalist MLLMs that unify vision-language understanding and editing.

\subsection{MLLM Privacy}
Prior works on privacy-preserving learning for MLLMs include Differential Privacy ($\text{DP}$)\cite{dp1,dp2,dualpriv} and inference-time protection methods like ReVision and MARRS\cite{revision,marrs}. While effective for general protection, these approaches fall short when tasks involve direct manipulation of visual content (e.g., image editing), as they lack concrete visual input. Surrogate-driven protection~\cite{surrogate1,surrogate2,surrogate3,chi} addresses this by replacing sensitive regions with synthetic content, but existing methods prioritize privacy over content consistency, often resulting in inconsistent outputs. Consequently, the crucial task of recovering intended outputs after editing has been largely overlooked—a significant challenge given the diverse and flexible nature of MLLM editing scenarios.

\section{SPPE Dataset}
\subsection{Dataset Overview}
\begin{table}[ht]
\raggedright
\renewcommand{\arraystretch}{1}

\setlength{\tabcolsep}{3pt}

\begin{tabular}{l|cccccr}
\toprule
% \hline
\textbf{Dataset} & 
\makecell[c]{\textbf{Anno.}} & 
\makecell[c]{\textbf{Prot.}} & 
\makecell[c]{\textbf{Edit.}} & 
\makecell[c]{\textbf{Rec.}} & 
\makecell[c]{\textbf{MLLM}} & 
\makecell[c]{\textbf{Scale}} \\
% \midrule
\hline
VISPR      &     &     &     &     &     &  12,000  \\
VizWiz-Priv & \ding{51} & \ding{51} &     &     &     &  5,537   \\
Redactions & \ding{51} &     &     &     &     & 8,473    \\
DIPA  & \ding{51} &     &     &     &     &   1,495  \\
DIPA2     & \ding{51} &     &     &     &     &   1,304  \\
BIV-Priv-Seg & \ding{51} &     &     &     &     &    728* \\
Multi-P2A  &     &     &     &     & \ding{51}  &   31,962 \\  
ReVision & & \ding{51} & & \ding{51} &\ding{51} &1,700 \\
HR-VISPR & &\ding{51} & && & 10,110\\

\hline

\textbf{Ours} & \ding{51} & \ding{51} & \makecell[c]{ \ding{51}(65)} & \ding{51} & \ding{51}  &    55,696 \\
\bottomrule

% \hline
\end{tabular}
\caption{Benchmark Comparison. \textbf{Anno.}: region-level private content annotation; \textbf{Prot.}: availability of protected versions; \textbf{Edit.}: suitability for image editing tasks; \textbf{Rec.}: recovery requirement; \textbf{MLLM}: relevance to MLLM use cases; Scale: dataset size.}
\label{tab:dataset}
\end{table}
\begin{figure*}[h]
    \centering
    \includegraphics[width=.9\linewidth]{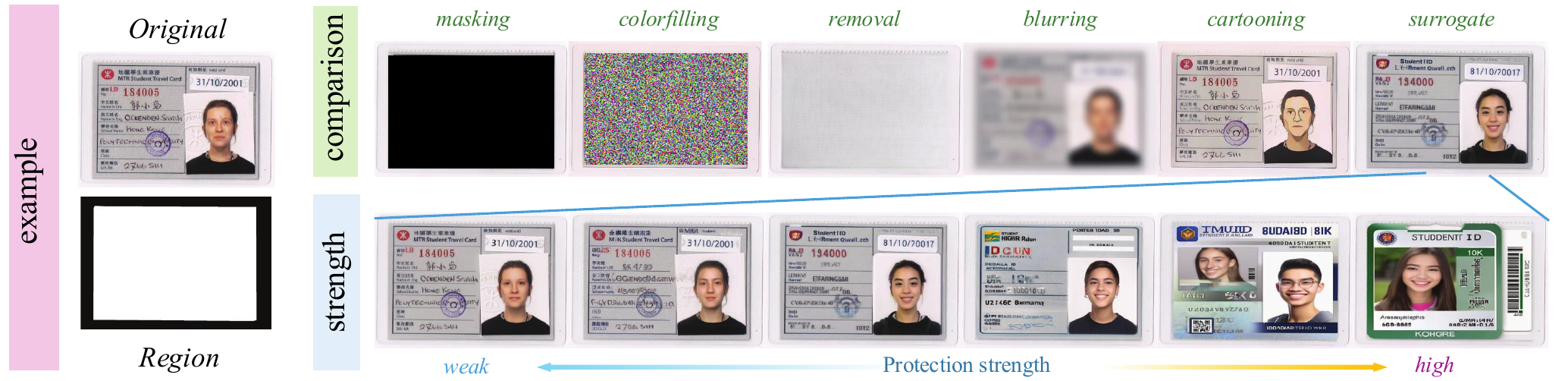}
\caption{Performance of surrogate generation. This example shows an image containing a sensitive student ID card region (leftmost panel). On the right, the top row compares our surrogate method with traditional privacy protection techniques, demonstrating superior concealment of private content while maintaining semantic coherence. The bottom row presents surrogates generated with varying protection strengths, where a higher strength indicates less influence from the original image. We select a middle strength that achieves a flexible trade-off between privacy protection and semantic fidelity.}

    \label{fig:guidance}
\end{figure*}

We propose SPPE, a new benchmark dataset specifically designed for evaluating the image editing quality of Multimodal Large Language Models (MLLMs) under surrogate-driven, privacy-preserving scenarios. 
Table~\ref{tab:dataset} compares SPPE with existing privacy protection datasets~\cite{vispr,dipa,vizwiz_priv,visual_redaction,dipa2,biv_priv,multip2a,revision,hrvispr}, showing that SPPE uniquely provides large-scale coverage and complete data support across protection, editing, and recovery stages, specifically tailored for evaluating privacy-aware MLLM editing.. Figure~\ref{fig:sample} illustrates a representative sample. Each instance in SPPE is represented as a 7-tuple: the sensitive category $C$, the edit instruction $P$, the original image $I$, the privacy mask $M$, the surrogate image $S$, the edited $S'$ and $I'$ obtained by applying prompt $P$ with MLLM.\\
SPPE is constructed through the following four stages: (1) Data Collection, (2) Surrogate Generation, (3) Prompt Definition, (4) MLLM Editing.

\paragraph{Data Collection}

\begin{figure}[!h]
\setlength{\textfloatsep}{1pt}
    \centering
    \includegraphics[width=.7\linewidth]{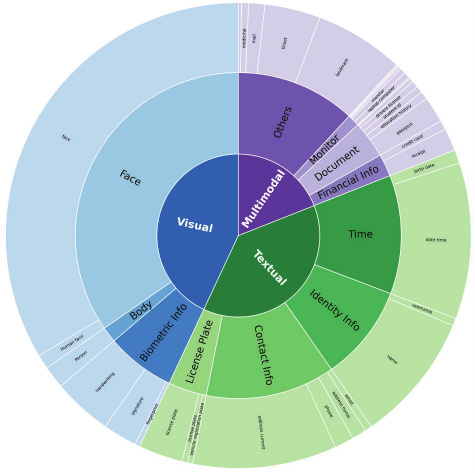}
    \caption{Distribution of privacy-related categories across three modality types (Textual, Visual, and Multimodal). The outer ring represents fine-grained categories, while the inner ring shows their grouping into higher-level modality classes.}
    \label{fig:data}
\end{figure}

We collect original images and object-level annotations from widely used privacy datasets, including VISPR~\cite{vispr}, Visual Redaction~\cite{visual_redaction}, DIPA~\cite{dipa}, and DIPA2~\cite{dipa2}. A total of 36 fine-grained categories are unified into 10 high-level privacy types and further grouped into three modalities—textual, visual, and multimodal—based on their appearance. Figure~\ref{fig:data} illustrates the distribution, showcasing broad coverage of privacy types to support realistic and robust evaluation.
\paragraph{Surrogate Generation}
We generate privacy-preserving surrogates by inpainting masked sensitive regions using the SDXL model~\cite{sdxl}. As shown in the top right of Figure~\ref{fig:guidance}, this approach preserves image utility better than traditional methods. The protection strength can be adjusted by varying the influence of the original image (see bottom right); we adopt a moderate guidance scale of 0.5 to balance privacy and utility.
\begin{figure*}
    \centering
    \includegraphics[width=1\linewidth]{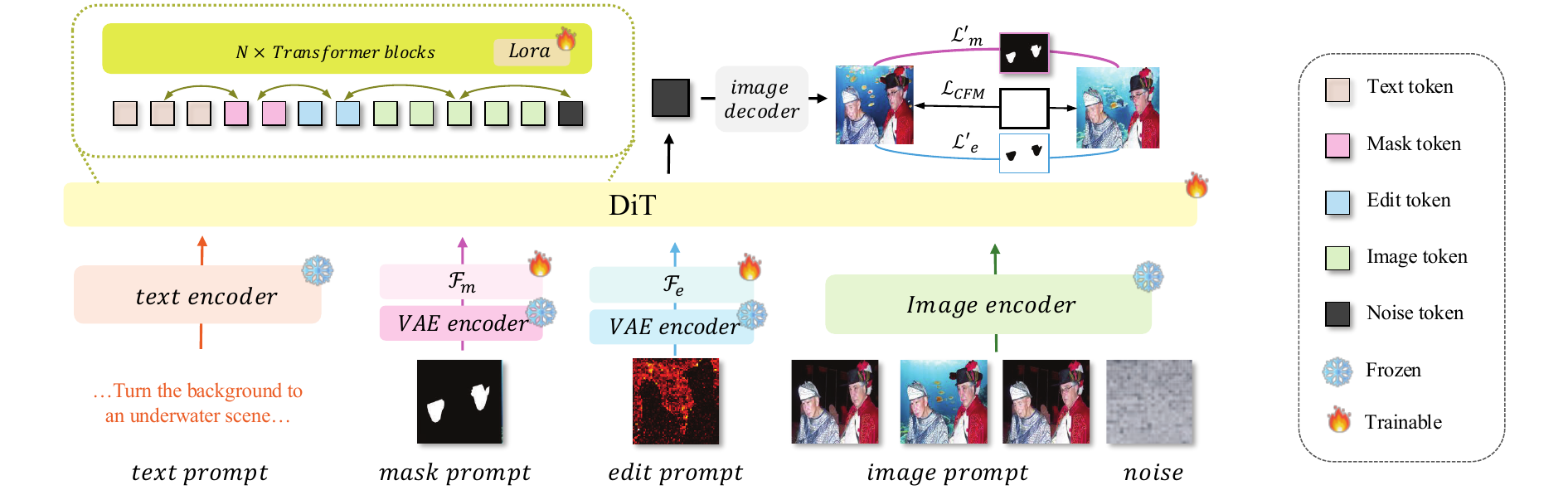}
    \caption{Overview of the proposed Surrogate-to-Original Editable Recovery (SOER). SOER processes semantic, visual, restoration, and edit cues through dedicated encoders to extract rich spatial and directional information. The resulting embeddings are combined and fed into a DiT-based transformer for multimodal interaction, enabling the generation of an edited image that faithfully reflects MLLM-driven edits while remaining consistent with the original content.}

    \label{fig:local_transfer}
\end{figure*}
\paragraph{Prompt definition}
To simulate flexible real-world user operations, we defined 65 prompts covering a variety of editing types, including style transfer, local feature modification, object addition or removal, etc. Some examples are shown in Table~\ref{tab:editing_types}, and the full set of edits can be found in the appendix.
\begin{table}[h]
\centering
\begin{tabular}{c|c}
% \hline
\toprule
\textbf{Type} & \textbf{Example} \\
\hline
Style & Turn to a pencil sketch style...\\
Concept & Turn the person into a vampire...\\
Addition & Let the person wear sunglasses...\\
Removal & Remove the text... \\
Replace & Change the background to a forest... \\
Appearance & Make the person look older,... \\
... & ...\\
% \hline
\bottomrule
\end{tabular}
\caption{Examples of editing prompts.}
\label{tab:editing_types}
\end{table}
\subsubsection{MLLM editing}

We employ SmartEdit~\cite{smartedit} to generate MLLM-style edits for both the original image $I$ and its surrogate $S$. Two prompts are randomly selected per instance to produce $(S'_1, I'_1)$ and $(S'_2, I'_2)$, where $I'_1$ and $I'_2$ serve as ground-truth references for evaluating the fidelity and accuracy of local recovery.

\section{Surrogate-to-Original Editable Recovery}

\subsection{Task Definition}
To address the challenge of privacy-preserving MLLM image editing, we define a new recovery task. This task aims to reconstruct the MLLM output $\hat{I}$ locally without uploading the original image $I$ to the MLLM. The primary objective is to ensure that the reconstructed image $\hat{I}$ closely matches the output generated by directly applying the MLLM to the original image. Formally, given a sample as defined in the dataset section, the goal is to develop a recovery model $\mathcal{R}$  that generates $\hat{I}$ by :
\[
\hat{I'} = \mathcal{R}(I, S, S', C, M, P), 
\]

The goal is to minimize the difference between $\hat{I'}$ and $I'$.

\subsection{Surrogate-to-Original Editable Recovery}

We propose \textbf{SOER}, a DiT-based generative framework for privacy-aware output recovery that preserves the MLLM editing effects. Our method integrates four types of guidance—semantic, visual, restore, and edit—to generate a reconstructed MLLM output. These guidance cues enable effective control over spatial localization and precise perception of editing intent, allowing for fine-grained and controllable reconstruction that closely approximates the original MLLM-edited result.

\paragraph{Semantic guidance.}
The user’s editing intent is encoded using a pretrained text encoder applied to the instruction prompt $p_{\text{text}}$ (e.g., “add glasses” or “change background to sunset”). This semantic embedding serves as a high-level guidance signal that directs the model to apply structural and stylistic changes aligned with the intended modification.

\paragraph{Visual Guidance.}
% \paragraph{Visual Guidance.}
To support faithful and privacy-compliant reconstruction, we utilize visual references that encode both content priors and editing context. The original image $I$ conveys detailed private visual content, while the surrogate $S$ and its edited version $S'$ capture the transformation behavior induced by the MLLM. These three images are processed by a shared VAE-based encoder to produce embeddings $p_{\text{I}}, p_{\text{S}}, p_{\text{S'}}$, providing the recovery model with cues about the source content and MLLM editing effects.

\paragraph{Restore \& Edit Guidance.}
To enhance localization accuracy and improve disentanglement between preservation and modification, we further introduce two spatial priors. The privacy mask $M$, defined by differences between $I$ and $S$, highlights regions where private content has been replaced and must be faithfully preserved. The edit map, computed from the difference between $S$ and $S'$, identifies areas modified by the MLLM. While such information is partially encoded in the visual embeddings, these explicit region-level signals offer more precise guidance on where to recover original content and where to apply editing transformations. Both $M$ and the edit map are encoded using a VAE encoder $\mathcal{E}$ and passed through guidance-specific Transformers $\mathcal{F}_m$ and $\mathcal{F}_e$. Their outputs are averaged via module $\mathcal{A}$ to obtain compact embeddings $p_m$ and $p_e$:
\begin{equation}
p_{m} = \mathcal{A}(\mathcal{F}_m(\mathcal{E}(M))), \quad
p_{e} = \mathcal{A}(\mathcal{F}_e(\mathcal{E}(|S' - S|^2)))
\end{equation}

\paragraph{Multimodal Control Generation} 
All guidance embeddings and the noisy latent representation are first projected into a shared embedding space and concatenated into a composite input sequence:
\begin{equation}
G = [p_{\text{e}}, p_{\text{m}}, p_{\text{text}}, p_{\text{s}}, p_{\text{s'}}, p_{\text{I}}, n_{\hat{I}}].
\end{equation}
Here, $n_{\hat{I}}$ represents the noisy latent token to be progressively denoised into the target image $\hat{I}$. The denoising process is performed by a DiT backbone enhanced with Multimodal Attention (MMA) modules, which enable dynamic interaction and information exchange among editing intent, content representation, and spatial localization cues. Formally, MMA computes:
\[
Q = G W_Q, \quad K = G W_K, \quad V = G W_V,
\]
\[
\mathrm{MMA}(G) = \mathrm{softmax}\left(\frac{Q K^\top}{\sqrt{d_k}}\right) V,
\]
where $W_Q, W_K, W_V \in \mathbb{R}^{d \times d_k}$ are learnable projection matrices, and $d_k$ is the dimension of the key vectors.

\paragraph{Loss Computation.} We optimize the local reconstruction module via a Conditional Flow Matching (CFM) loss:

\begin{equation}
\mathcal{L}_{\text{all}} = \mathbb{E}_{t,\, p_t(x|z),\, q(z)} \left\| v_{\theta}(x, t) - u_t(x|z) \right\|^2
\end{equation}

% \subsection{Dual Training Strategy.} 

To jointly optimize both guidance pathways, we adopt an alternating training strategy: at each step, gradients are propagated through either the mask encoder $\mathcal{F}_m$ or the edit encoder $\mathcal{F}_e$, while freezing the other. Correspondingly, we apply region-aware supervision by re-weighting $\mathcal{L}_{\text{all}}$ based on the binary mask $M$:
\begin{equation}
\mathcal{L}_m = \mathcal{L}_{\text{all}} + \mathcal{L}_{\text{all}} \odot M, \quad
\mathcal{L}_e = \mathcal{L}_{\text{all}}  +\mathcal{L}_{\text{all}} \odot (1 - M)
\end{equation}
This strategy guides the model to maintain original content consistency in sensitive regions while ensuring faithful edit reproduction elsewhere. By applying region-weighted loss and alternately training the mask and edit encoders, the model learns to balance preservation and editing, achieving stable and effective guidance.

\begin{figure*}[h]
    \centering
    \includegraphics[width=.95\linewidth]{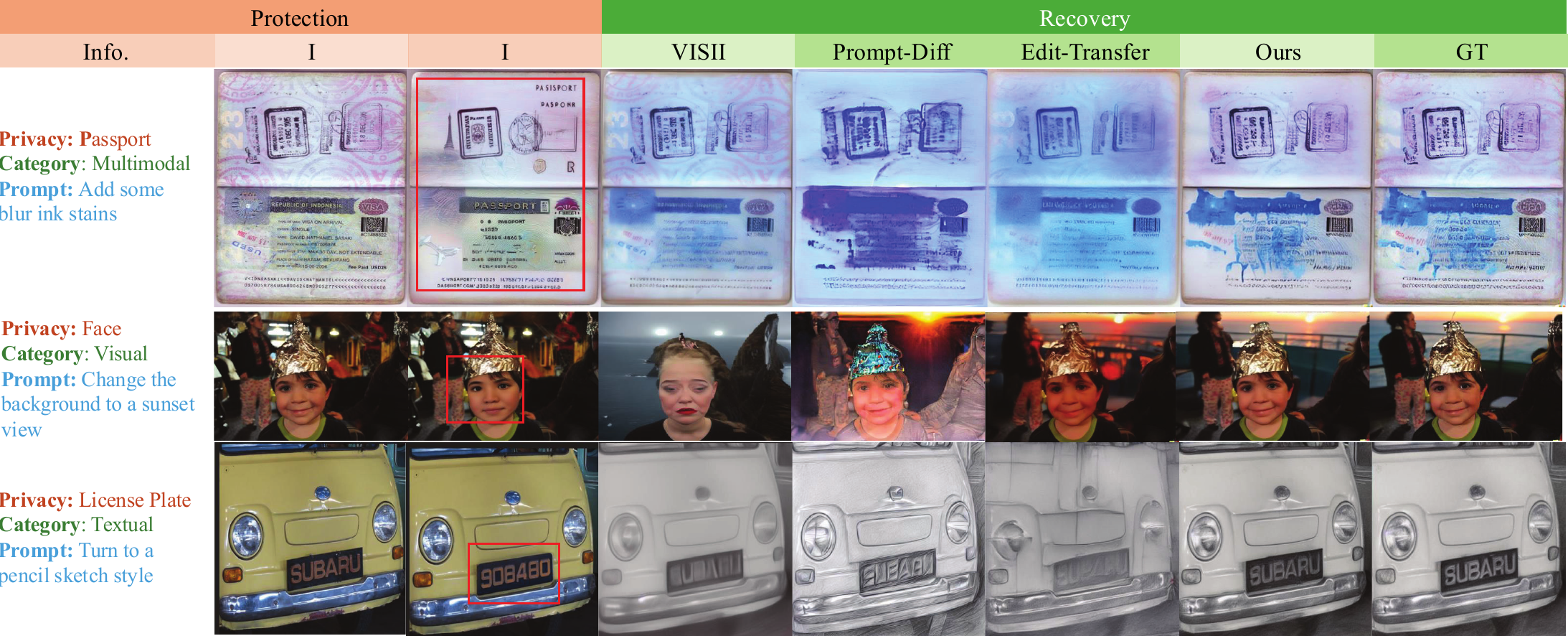}
    \caption{Qualitative comparison across textual, visual, and multimodal privacy scenarios. The examples highlight our method’s ability to accurately reproduce MLLM-driven edits, maintain semantic consistency with the original prompts, and effectively recover sensitive content in the source image. }
    \label{fig:enter-label}
\end{figure*}

\begin{table}[t]
\small
\centering

% \caption{Quantitative comparison on \textbf{SPPE} Dataset. We evaluate the \textit{Edit Consistency}  of all methods across multiple privacy categories. We compute SSIM, PSNR, and CLIP similarity to assess the edit effect similarity and two direction similarity DirS and DirI  further assess the alignment of image variation directions.}
% \setlength{\tabcolsep}{2pt} % Default value: 6pt
% \renewcommand{\arraystretch}{0.9} % Default value: 1
\renewcommand{\arraystretch}{1.1}
\begin{tabular}{cccccc}
% \toprule
\multicolumn{6}{c}{\textbf{TEXTUAL}} \\
% \multirow{2}{*}{\textbf{Method}} & \multicolumn{5}{c}{\textbf{Edit Transfer}}  \\
% \hline
\toprule

\textbf{Method}& \textbf{SSIM}$\uparrow$  & \textbf{PSNR} $\uparrow$ & \textbf{CSIM}$\uparrow$ &\textbf{DirS }$\uparrow$ &\textbf{DirI }$\uparrow$ \\
\hline
\textbf{GT} &1.0000  & inf& 1.0000 & 0.7571&1.0000 \\
\cdashline{1-6}[1pt/1pt]
% \rowcolor{black!10}
\textbf{P-diff} &0.4437 & 14.4026&  0.7031&0.5605 & 0.6269 \\
\textbf{VISII}&0.6778&20.3235 & 0.7567& 0.5435&0.6245\\
\textbf{E-Trans}& \underline{0.7483}& \underline{20.6851}& \underline{0.8126}& \underline{0.6717}&\underline{0.7223 }\\
% \rowcolor{yellow!20}
\rowcolor{black!10}
\textbf{Ours} & \textbf{0.8049}& \textbf{22.9952}&\textbf{0.8482}& \textbf{0.7139} & \textbf{0.7584}\\
% \bottomrule

% \begin{tabular}{c|ccccc}
% \toprule
\bottomrule
\multicolumn{6}{c}{\textbf{MULTIMODAL}} \\
% \multirow{2}{*}{\textbf{Method}} & \multicolumn{5}{c}{\textbf{Metrics}}  \\
% \hline
\toprule
\textbf{Method}& \textbf{SSIM}$\uparrow$  & \textbf{PSNR} $\uparrow$ & \textbf{CSIM}$\uparrow$ &\textbf{ DirS }$\uparrow$ &\textbf{DirI }$\uparrow$ \\
\hline
\textbf{GT} &1.0000 & inf& 1.0000  & 0.5899&1.0000 \\
\cdashline{1-6}[1pt/1pt]
% \rowcolor{black!10}
\textbf{P-diff} &0.4400& 14.1876&  0.7124&0.4541 & 0.6145 \\
\textbf{VISII}&0.6669&19.5801 &  0.7535& 0.4181&0.6025\\
\textbf{E-Trans}& \underline{0.7144}& \underline{19.6907}& \underline{0.7961}& \underline{0.5283}& \underline{0.6867}\\
% \rowcolor{yellow!20}
\rowcolor{black!10}
\textbf{Ours} & \textbf{0.7608}& \textbf{21.6246}&\textbf{0.8267}& \textbf{0.5631}& \textbf{0.7195}\\
\bottomrule

\multicolumn{6}{c}{\textbf{VISUAL}} \\ 
% \multirow{2}{*}{\textbf{Method}} & \multicolumn{5}{c}{\textbf{Metrics}}  \\
\toprule
\textbf{Method}& \textbf{SSIM}$\uparrow$  & \textbf{PSNR} $\uparrow$ & \textbf{CSIM}$\uparrow$ &\textbf{ DirS }$\uparrow$ &\textbf{DirI }$\uparrow$ \\
\hline
\textbf{GT} &1.0000  &inf & 1.0000 & 0.7281&1.0000 \\
\cdashline{1-6}[1pt/1pt]
% \rowcolor{black!20}
\textbf{P-diff} &0.5350 & 15.2398& 0.7614&0.5360 & 0.6114 \\
\textbf{VISII }&0.5715 &16.5407 & 0.7292& 0.4360&0.5270\\
\textbf{E-Trans }& \underline{0.7600}& \underline{20.4163}& \underline{0.8350}& \underline{0.6143}&\underline{0.6828} \\
% \rowcolor{yellow!20}

\rowcolor{black!10}
\textbf{Ours} & \textbf{0.8166}&\textbf{22.9206}& \textbf{0.8693}& \textbf{0.6656}& \textbf{0.7274}\\
\bottomrule
\end{tabular}
% \caption*{\footnotesize \textit{Note:} VISII is not included in this comparison since it uses a model pretrained on InstructPix2Pix during inference, giving it an unfair advantage in this specific evaluation.}
\caption{\textit{Edit Consistency.} Evaluations on the SPPE dataset demonstrate that our SOER model consistently outperforms existing methods, including VISII, P-Diff, and E-Transfer. The best and second-best results are highlighted in \textbf{bold} and \underline{underlined}, respectively. Arrows ($\uparrow$) indicate that higher values are preferable. SOER achieves best performance across all metrics, showcasing its superior ability to preserve MLLM-intended edits under privacy-preserving conditions.
}
\label{tab:category_reproduction_results}
\end{table}

\section{Experiment}
\paragraph{Datasets.}

We evaluate our method on both the proposed \textbf{SPPE} dataset and the public \textbf{InstructPix2Pix} dataset. For fair comparison, all methods (VISII~\cite{visii}, Prompt-Diffusion~\cite{prompt-diffusion}, and Edit-Transfer~\cite{edittransfer}) are trained on the SPPE training split. Evaluation is conducted on the SPPE test set and a subset of the InstructPix2Pix dataset, covering 934 unique prompts. Since InstructPix2Pix lacks sensitive region annotations, we generate full-image surrogates for its samples. While InstructPix2Pix contains synthetic images and prompts absent from SPPE, it offers a comprehensive benchmark for assessing model generalization to real-world scenarios where users may provide novel edit instructions.

\begin{table}[!h]
\small
\centering
\setlength{\tabcolsep}{3pt} % Default value: 6pt

\renewcommand{\arraystretch}{1.1}

\begin{tabular}{cccc}
\multicolumn{4}{c}{\textbf{TEXTUAL}}  \\
\toprule
\textbf{Method} & \textbf{SSIM} & \textbf{PSNR} & \textbf{CSIM} \\
\hline
\textbf{GT} & 0.6271 & 16.4797 & 0.6514 \\
\cdashline{1-4}[1pt/1pt]
\textbf{P-diff} & 0.3750 (-0.2521) & 11.4191 (-5.0606) & 0.5069 (-0.1445) \\
\textbf{VISII} & 0.6891 (+0.0620) & 17.4144 (+0.9347) & 0.6800 (+0.0286) \\
\textbf{E-Trans} & 0.6150 (\underline{-0.0121}) & 16.0001 (\underline{-0.4796}) & 0.6161 (\underline{-0.0353}) \\
\rowcolor{black!10}
\textbf{Ours} & 0.6358 (\textbf{+0.0087}) & 16.4096 (\textbf{-0.0701}) & 0.6456 (\textbf{-0.0058}) \\
\bottomrule

% \begin{tabular}{cccc}
\multicolumn{4}{c}{\textbf{MULTIMODAL}}  \\
\toprule
\textbf{Method} & \textbf{SSIM} & \textbf{PSNR} & \textbf{CSIM} \\
\hline
\textbf{GT} & 0.6310 & 16.1426 & 0.6706 \\
\cdashline{1-4}[1pt/1pt]
\textbf{P-diff} & 0.3860 (-0.2450) & 11.3131 (-4.8295) & 0.5401 (-0.1305) \\
\textbf{VISII} & 0.6864 (+0.0554) & 17.3335 (+1.1909) & 0.6966 (\underline{+0.0260}) \\
\textbf{E-Trans} & 0.6160 (\underline{-0.0150}) & 15.7191 (\underline{-0.4235}) & 0.6359 (-0.0347) \\
\rowcolor{black!10}
\textbf{Ours} & 0.6372 (\textbf{+0.0062}) & 16.0705 (\textbf{-0.0721}) & 0.6644 (\textbf{-0.0062}) \\
\bottomrule
% \end{tabular}

% \begin{tabular}{cccc}
\multicolumn{4}{c}{\textbf{VISUAL}}  \\
\toprule
\textbf{Method} & \textbf{SSIM} & \textbf{PSNR} & \textbf{CSIM} \\
\hline
\textbf{GT} & 0.6499 & 16.4889 & 0.7401 \\
\cdashline{1-4}[1pt/1pt]
\textbf{P-diff} & 0.4568 (-0.1931) & 12.2344 (-4.2545) & 0.6214 (-0.1187) \\
\textbf{VISII} & 0.6084 (-0.0415) & 16.3341 (\underline{-0.1548}) & 0.6829 (-0.0572) \\
\textbf{E-Trans} & 0.6368 (\underline{-0.0131}) & 16.0187 (-0.4702) & 0.7098 (\underline{-0.0303}) \\
\rowcolor{black!10}
\textbf{Ours} & 0.6563 (\textbf{+0.0064}) & 16.4994 (\textbf{+0.0105}) & 0.7328 (\textbf{-0.0073}) \\
\bottomrule
% \end{tabular}

\end{tabular}

\caption{
\textit{Source Integrity} Evaluations on the SPPE dataset show that our SOER model consistently achieves results closest to the ground-truth (GT) reference. The best and second-best results are marked in \textbf{bold} and \underline{underlined}.  
}
\label{tab:preservation}
\end{table}
\noindent \textbf{Evaluation Metrics.}
We evaluate privacy-aware MLLM editing from two perspectives: \textit{Edit Consistency} and \textit{Source Integrity}. \textit{Edit Consistency} measures how well the edited image $\hat{I}$ aligns with the intended edits $I'$ by the MLLM, using metrics such as CLIP Similarity (CLIP-Sim), Structural Similarity Index (SSIM), Peak Signal-to-Noise Ratio (PSNR), and Directional Similarity (DirS/DirI) \cite{visii, instanip}, which compares editing directions between image pairs like ($I \rightarrow \hat{I}$) and ($S \rightarrow S'$) or ($I \rightarrow I'$). \textit{Source Integrity} evaluates how well $\hat{I}$ preserves the original content by comparing it with $I$ using CLIP-Sim, SSIM, and PSNR. Since minimizing edits can artificially inflate these metrics, Source Integrity is considered jointly with edit-related metrics for balanced evaluation. To this end, we use the corresponding metric values computed on the ground-truth edited image $I'$ as a reference to assess how closely $\hat{I}$ matches the expected editing outcomes.

\noindent \textbf{Training Setup.}
Our method builds on the FLUX Inpainting model, which adopts a DiT-based architecture with a T5 text encoder and a VAE image encoder. We fine-tune the model on the SPPE training split using LoRA (rank 256) for 6000 steps, optimizing with 8-bit AdamW at a learning rate of $1 \times 10^{-4}$. Inference is performed with 35 denoising steps.

\subsection{Qualitative Results}
Fig.~\ref{tab:category_reproduction_results} presents qualitative results across various privacy scenarios (visual, textual, and multimodal), demonstrating three key strengths of our method. First, we achieve superior \textit{instruction fidelity}, as seen in the top example where our approach accurately realizes the intended modification (e.g., adding blue ink stains), in contrast to other methods that produce incomplete or imprecise edits. Second, we achieve superior \textit{MLLM consistency}. In the middle example, although existing methods like Prompt-Diff and Edit-Transfer successfully perform the requested edit (e.g., changing the background to a sunset view), our output aligns more faithfully with the actual MLLM's style and details. Finally, the bottom example highlights our robust \textit{privacy preservation}. Despite global transformations such as style transfer, our model successfully disentangles edited regions from areas that must remain consistent with the original image, thereby accurately simulating MLLM edits on the original content while retaining sensitive details (e.g., the license plate).

\subsection{Quantitative Results}

\begin{table}[h]
\small
\centering
\renewcommand{\arraystretch}{1.1}

\begin{tabular}{cccccc}
% \toprule
\multicolumn{6}{c}{\textbf{EDIT CONSISTENCY}} \\
% \multirow{2}{*}{\textbf{Method}} & \multicolumn{5}{c}{\textbf{Edit Transfer}}  \\
\toprule

\textbf{Method}& \textbf{SSIM}$\uparrow$  & \textbf{PSNR} $\uparrow$ & \textbf{CSIM}$\uparrow$ &\textbf{DirS }$\uparrow$ &\textbf{DirI }$\uparrow$ \\
\hline
% \rowcolor{black!10}
\textbf{GT} &1.0000 & inf &1.0000 & 0.2705&1.0000\\
\cdashline{1-4}[1pt/1pt]
\textbf{P-diff} &0.4111&12.2141&  0.7417&0.1568& 0.3699 \\
\textbf{E-Trans}&   \underline{0.5251}& \underline{15.9315}& \underline{0.7661}& \underline{0.1717}& \underline{0.3446} \\
\rowcolor{black!10}
\textbf{Ours} &\textbf{ 0.5340}& \textbf{16.0415}&\textbf{ 0.7857}& \textbf{0.2077}& \textbf{0.3817} \\
\bottomrule
\end{tabular}

% \end{tabular}
\setlength{\tabcolsep}{3pt}
\begin{tabular}{cccc}
\multicolumn{4}{c}{\textbf{SOURCE INTEGRITY}}  \\
\toprule
 Method  & SSIM   &  PSNR &  CSIM\\
\hline
\textbf{GT} & 0.6914 & 17.7684 & 0.8769 \\
\cdashline{1-4}[1pt/1pt]
\textbf{P-diff} & 0.4300 (-0.2614) & 11.7821 (-5.9863) & 0.7248 (-0.1521) \\
\textbf{E-Trans}  & 0.5320 (\underline{-0.1594}) & 15.1055 (\underline{-2.6629}) & 0.7684 (\underline{-0.1085}) \\
\rowcolor{black!10}
\textbf{Ours}         & 0.5413 (\textbf{-0.1501}) & 15.1643 (\textbf{-2.6041}) & 0.7783 (\textbf{-0.0986}) \\
\bottomrule
\end{tabular}
\caption{Quantitative comparison on the \textbf{InstructP2P} dataset. We adopt similar evaluation protocols to measure \textit{Edit Consistency} and \textit{Source Integrity}, assessing the model's generalization ability on unseen images and prompts.}

\caption*{\footnotesize \textit{Note:} VISII is not included in this comparison since it uses a model pretrained on InstructP2P during inference, giving it an unfair advantage in this specific evaluation.}
\label{tab:ip2p_res}
\end{table}

Our method achieves the highest Edit Consistency across a wide range of privacy categories and edit types, significantly outperforming baselines and consistently producing metrics that are closest to the ground truth. For instance, in the Multimodal category, our method not only surpasses all competitors with a DirS of 0.5631 but also achieves the smallest divergence from the GT value 0.5899, demonstrating faithful capturing of MLLM edit hints.
Our approach also consistently yields Source Integrity scores that are closest to the ground truth, striking a strong balance between faithful editing and content preservation. For example, while methods like VISII may occasionally score highest in Source Integrity (e.g., an SSIM of 0.6891 in the Textual category compared to our 0.6358), they often exhibit poor Edit Consistency (with an SSIM of only 0.6778 against our 0.8049), suggesting a tendency toward under-editing. These results highlight the advantage of our approach in achieving privacy-aware edits that are both semantically aligned and visually coherent.\\
Table~\ref{tab:ip2p_res} presents results on the Instruct-p2p dataset, which requires generalization across unseen prompts and a full-image surrogate. The use of a full-image surrogate naturally introduces greater discrepancies with the original image, which in turn makes the environment more challenging and leads to a notably low ground-truth alignment (DirS of only 0.2705).
Despite these challenging conditions, our approach demonstrates robust performance. Our method achieves particularly strong results on the DirS (0.2077) and DirI (0.3817) metrics, significantly outperforming competitors. The demonstrated robustness highlights our approach's potential as a privacy-preserving editing solution that can be effectively integrated into diverse real-world applications with minimal need for costly retraining.

 \subsection{Ablation study}

\begin{figure}
    \centering
    \includegraphics[width=.9\linewidth]{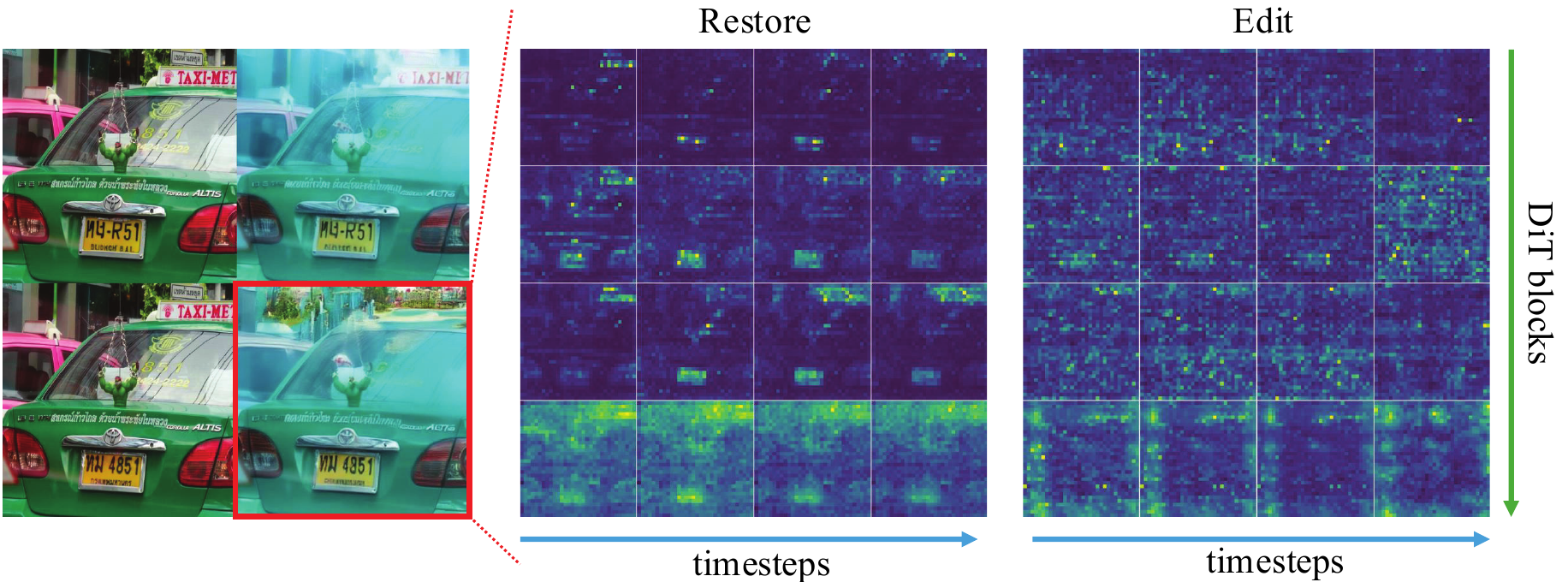}
    \caption{We visualize attention maps corresponding to the restore and edit guidance tokens, revealing distinct attention patterns—restore tokens focus on sensitive regions requiring preservation (e.g., license plates), while edit tokens highlight areas of semantic modification (background), reflecting the spatial roles of the two guidance signals.
}
    \label{fig:attention}
\end{figure}
\begin{table}[ht]
\small
\centering

\renewcommand{\arraystretch}{1.1}

\begin{tabular}{cccccc}
\toprule
& \textbf{SSIM}$\uparrow$  & \textbf{PSNR} $\uparrow$ & \textbf{CSIM}$\uparrow$ &\textbf{DirS }$\uparrow$ &\textbf{DirI }$\uparrow$ \\
\hline
\rowcolor{black!10}
\textbf{Ours} & 0.8016& 22.7111&0.8531& 0.6658& 0.7382\\
% \hline
w/o mask &  0.7892 & 21.9887& 0.8431 & 0.6498 &0.7239 \\
w/o diff & 0.7867 &21.9061 & 0.8433 &0.6511 & 0.7258\\
w/o text & 0.7139 &19.1110 & 0.8135 & 0.6309&  0.7070 \\
\hline
- CNN & 0.7178 &19.7713 & 0.7925 &0.5924 & 0.6725\\
% - Avg & 0. & & 0. & & \\
- CLS token & 0.7703 & 21.7016& 0.8270 &0.6196 & 0.7024\\
\bottomrule
\end{tabular}
\caption{Ablation study on multimodal guidance components and encoding strategies. We evaluate the impact of removing individual guidance signals, excluding the no-visual-guidance setting due to its severe degradation. We also compare different encoders for mask and edit map representations, where the VAE-based encoder demonstrates superior to CNN based and CLS-bsed methods.}
\label{tab:ablation_results}
\end{table}

% \subsection{Visualization}
\noindent \textbf{Impact of Multimodal Guidance Signals.}
We conduct an ablation study (Table~\ref{tab:ablation_results}) to evaluate the contribution of each guidance component. Removing the text prompt leads to significant degradation, highlighting the importance of semantic alignment with user intent. Excluding spatial cues (e.g., mask or edit prompt) also results in performance drops, confirming their role in localizing edits. We further examine the spatial influence patterns in the Visualization section. These results demonstrate the complementary nature of all guidance signals in supporting accurate generation. The no-visual-guidance setting is excluded due to performance collapse caused by the lack of reference inputs.

\noindent \textbf{Representation Strategies for Spatial Guidance.}
We assess the impact of different representation strategies for restore and edit guidance. Substituting our transformer-based encoder with a CNN, or summarizing spatial inputs using a single CLS token leads to notable performance drops, highlighting the importance of expressive, context-aware encoders in capturing edit intent and spatial cues.

\subsection{Visualization}
To better understand how restore and edit guidance signals influence generation, we visualize attention maps across DiT blocks and denoising steps (Figure~\ref{fig:attention}). The restore guidance exhibits focused attention on privacy-sensitive regions (e.g., license plates), facilitating accurate content preservation. In contrast, the edit guidance attends to regions aligned with the intended visual effect—for example, emphasizing a more global focus in this case, where the prompt requires transforming the scene into an underwater style.

\section{Conclusion}

This work focused on a critical yet overlooked challenge: enabling MLLM image editing under surrogate-driven privacy protection. We introduced SPPE, a novel benchmark for evaluating edit fidelity in privacy-aware settings, and proposed a multimodal generative framework, SOER, that locally recovered edited images consistent with both the original content and MLLM edits. Extensive experiments demonstrated that SOER outperformed existing methods in simulating MLLM effects while preserving private content, even in challenging scenarios. Our work established a new standard in the field and provided a practical solution for real-world MLLM applications under privacy constraints.

\section{Acknowledgement}

This work acknowledges the Hon Hai-CityU Joint Research Center and Hon Hai Research Institute for their financial and technical support.

\bibliography{aaai2026}
\end{document}